\pdfoutput=1

\documentclass[11pt]{article}

\usepackage[final]{acl}

\usepackage{times}
\usepackage{latexsym}
\usepackage{amsmath}
\usepackage{bbm}
\usepackage{verbatim}
\usepackage{booktabs}
\usepackage{multirow}

\usepackage[T1]{fontenc}

\usepackage[utf8]{inputenc}

\usepackage{microtype}

\usepackage{inconsolata}

\usepackage{graphicx}

\newcommand\blfootnote[1]{%
  \begingroup
  \renewcommand\thefootnote{}\footnote{#1}%
  \addtocounter{footnote}{-1}%
  \endgroup
}

\title{Toward Reliable Clinical Coding with Language Models: \\ Verification and Lightweight Adaptation}

\author{Zhangdie Yuan*$^{\dagger}$ \\
  University of Cambridge \\
  \texttt{zy317@cam.ac.uk} \\ \And
  Han-Chin Shing* \\
  Amazon  \\
  \texttt{hanchins@amazon.com} \\ \AND
    Mitch Strong \\
  Amazon  \\
  \texttt{mjstrong@amazon.com} \\ \And
    Chaitanya Shivade \\
  Amazon  \\
  \texttt{shivadc@amazon.com} \\}

\begin{document}
\maketitle

\begin{abstract}
Accurate clinical coding is essential for healthcare documentation, billing, and decision-making. While prior work shows that off-the-shelf LLMs struggle with this task, evaluations based on exact match metrics often overlook errors where predicted codes are hierarchically close but incorrect. Our analysis reveals that such hierarchical misalignments account for a substantial portion of LLM failures. We show that lightweight interventions, including prompt engineering and small-scale fine-tuning, can improve accuracy without the computational overhead of search-based methods. To address hierarchically near-miss errors, we introduce clinical code verification as both a standalone task and a pipeline component. To mitigate the limitations in existing datasets, such as incomplete evidence and inpatient bias in MIMIC, we release an expert double-annotated benchmark of outpatient clinical notes with ICD-10 codes. Our results highlight verification as an effective and reliable step toward improving LLM-based medical coding.\blfootnote{$^{*}$ Equal contribution.}
\blfootnote{$^{\dagger}$ Work done during internship at AWS AI.}\end{abstract}

\section{Introduction}

Accurate clinical coding (often used interchangeably with medical coding) is essential for healthcare documentation, billing, and decision-making \citep{johnson2016mimic, shickel2017deep}. Standardized coding systems such as the International Classification of Diseases, Tenth Revision, Clinical Modification (ICD-10-CM)\footnote{\url{https://www.cdc.gov/nchs/icd/icd-10-cm}} ensure consistency across medical records \citep{hirsch2016icd, deyoung2022entity}.
However, assigning correct codes to clinical notes remains highly challenging due to variability in medical narratives and the hierarchical complexity of ICD-10-CM, where only leaf nodes are valid for billing \citep{jha2009use}. The task requires selecting up to 12 correct codes from a set of 72,000 billable ICD-10-CM codes, making it a highly complex classification problem.

\begin{figure}[t]
    \centering
    \includegraphics[width=\columnwidth]{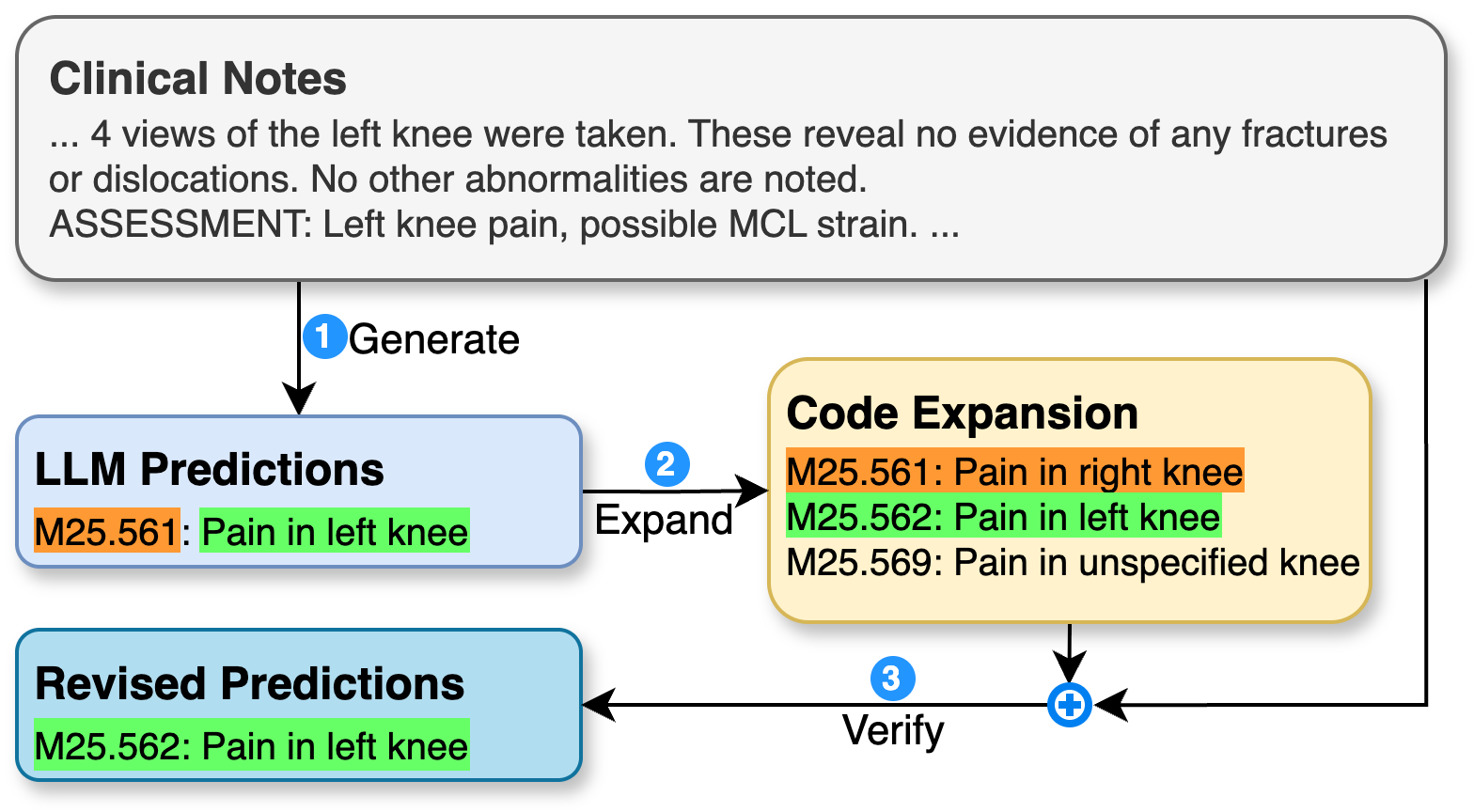} 
    \caption{An illustration of our generate-expand-verify pipeline. In this obfuscated example, the model-predicted code has the correct description with the wrong ICD-10-CM code. The expansion step uses ICD-10-CM tabular table to lookup its siblings. The verification step then selects the correct code and description based on the clinical notes and the expansion candidates.}
    \label{fig:graph}
\end{figure}

LLMs has spurred interest in automating medical coding. However, prior research has shown that off-the-shelf LLMs perform poorly, with low scores on exact match metrics such as F1 \citep{boyle2023automated, soroush2024large}. These findings align with our extensive evaluation of models spanning various families, sizes, and architectures. Scaling up model size alone does not solve the problem, and a significant portion of the errors are near misses, where predicted codes are hierarchically close but not exact matches. For example, an LLM may generate \texttt{I11.0} (Hypertensive heart disease with heart failure) instead of \texttt{I11.9} (Hypertensive heart disease without heart failure). Such errors expose a fundamental limitation in current evaluation approaches, which rely on exact match metrics and fail to capture hierarchical relationships. Prior work~\citep{liu2022hierarchical} incorporated ICD hierarchies via specialized loss functions, but focused on pre-LLM models, lacking its effect on LLMs.

Also, existing datasets such as MIMIC-III/IV \citep{johnson2016mimic} present challenges for the medical coding task. Not only do they use ICD-9 instead of ICD-10 codes, but the data is also limited to notes from the ICU, while the codes span all departments for the entire encounter~\citep{adams-etal-2022-learning}. This makes it difficult to determine whether a code is verifiable based solely on the note, limiting their suitability for our task, where explicit evidence in notes is critical. Moreover, both datasets focus primarily on inpatient records, whereas our work targets outpatient settings, where notes tend to be shorter, less structured, and more contextually straightforward \citep{shing2021clinicalencountersummarizationlearning}.

To address these limitations, we first construct a new double expert-annotated benchmark of outpatient clinical notes with ICD-10-CM codes, based on ACI-Bench~\citep{yim2023aci}.\footnote{The dataset accompanying this work is publicly  available at \url{https://github.com/amazon-science/toward-clinical-coding-verification-adaptation}.
}
Each note is double-annotated and adjudicated to ensure high-quality supervision. Then we investigate two key research questions. The first is understanding the extent to which off-the-shelf LLMs can perform clinical coding and the nature of the errors they make. A systematic error analysis reveals that hierarchical misalignments account for a substantial portion of mistakes, emphasizing the need for evaluation methods that go beyond exact match. The second question explores whether lightweight interventions, such as prompt engineering and small-scale fine-tuning, can significantly improve LLM performance.
These approaches offer practical solutions that are cost-effective and adaptable to evolving medical coding standards. To further address these challenges, we introduce a clinical code verification pipeline (\autoref{fig:graph}) that refines LLM predictions by leveraging the hierarchical structure of ICD-10-CM. The pipeline first expands candidate codes using ICD relationships and then refines these predictions using LLM-based verification to select the most contextually appropriate code. This verification step mitigates hierarchical misalignments and improves overall accuracy, up to 16 F1 for Haiku-3, for more reliable clinical coding.

\section{Improving LLMs for Clinical Coding}

\subsection{Prompt Engineering} \label{prompt_eng}
Prompt engineering is a lightweight, model-agnostic method to adapt LLMs without additional training data. To investigate how structured prompts influence clinical coding performance, we evaluate five prompt types: a single-line baseline, detailed instructions, chain-of-thought (CoT) reasoning, prompt decomposition (identifying key clinical phrases before prediction), and a combination of detailed instructions with CoT. The single-line prompt follows prior work~\citep{boyle2023automated} and is shown in~\autoref{sec:single_line_prompt}. We also incorporate the \textit{MEAT (Monitor, Evaluate, Assess, Treat) principle} into detailed instructions to provide structured clinical context, mimicking expert reasoning. By explicitly including descriptions and structured reasoning, we hypothesize that LLMs can better align predictions with the ICD-10-CM hierarchy.

\subsection{Fine-Tuning}
Fine-tuning enables task-specific adaptation of LLMs by training them on small, high-quality datasets. We investigate variants of fine-tuning to evaluate how different combinations of codes and descriptions affect performance. Specifically, we fine-tune the models to generate: 
1) codes alone, 2) descriptions alone, 3) codes followed by descriptions, and 4) descriptions followed by codes. These configurations correspond directly to the experiments in~\autoref{sec:results}, with results shown in ~\autoref{tab:fine_tune_results}.

Formally, let \( N = \{n_1, n_2, \ldots, n_m\} \) represent a set of clinical notes paired with gold-standard codes \( C^* = \{C^*_1, C^*_2, \ldots, C^*_m\} \). The model is trained to minimize the cross-entropy loss:
\[
\mathcal{L}_{\text{fine-tune}} = - \frac{1}{m} \sum_{i=1}^m \log P(C^*_i \mid n_i),
\]
where \( P(C^*_i \mid n_i) \) denotes the model’s predicted probability of the correct code \( C^*_i \) given the note \( n_i \). These configurations allow us to systematically evaluate the effects of different task-specific data representations. The goal is to provide a lightweight yet effective alternative to compute-intensive methods, such as retrieval-augmented generation, while aligning with the structured nature of ICD-10-CM.

\section{Clinical Code Verification}

\subsection{Overview of ICD-10-CM Structure} \label{structure}
Verification is critical in real-world healthcare systems, where erroneous codes can lead to financial, legal, and clinical consequences. We first introduce the structure of ICD-10-CM, which is a hierarchical coding system organized into two official lists: the \textit{tabular list} and the \textit{index list}. The tabular list is a tree structure where nodes represent ICD-10-CM codes, and edges encode parent-child relationships. Only leaf nodes, referred to as \textit{billable codes}, are valid for billing purposes. Formally, the tabular list is represented as a tree \( \mathcal{T} = (V, E) \), where \( V \) is the set of all ICD-10-CM codes and \( E \subseteq V \times V \) encodes the parent-child relationships.

For a code \( c \in V \), its parent is defined as \( P(c) = \{p \in V : (p, c) \in E\} \). The set of siblings, \( S(c) \), consists of codes that share the same parent as \( c \), formally \( S(c) = \{s \in V : P(s) = P(c) \text{ and } s \neq c\} \). Similarly, the set of cousins, \( C(c) \), includes codes that share the same grandparent as \( c \), defined as \( C(c) = \{g \in V : P(P(g)) = P(P(c)) \text{ and } g \notin S(c)\} \), shown in Appendix~\ref{appendix:icd_structure}. 

The index list, on the other hand, is represented as an undirected graph \( \mathcal{G} = (V, E') \), where the edges \( E' \subseteq V \times V \) encode cross-references between codes based on textual similarity or alternative naming conventions. For a code \( c \in V \), its 1-hop neighbors are defined as \( N_1(c) = \{n \in V : (c, n) \in E'\} \), and its 2-hop neighbors as \( N_2(c) = \{n \in V : \exists v \in V, (c, v) \in E' \text{ and } (v, n) \in E'\} \).

\subsection{Clinical Code Verification Task Definition}
Clinical code verification aims to determine whether a pre-assigned set of candidate codes \( \hat{C} = \{\hat{c}_1, \ldots, \hat{c}_k\} \) accurately reflects the information in a given clinical note \( N \). Each candidate \( \hat{c}_i \) is assigned a binary decision \( y_i \in \{0, 1\} \), where \( y_i = 1 \) indicates that \( \hat{c}_i \) matches a gold-standard code \( C^* \). Formally, the task is to produce binary labels \( \{y_1, y_2, \ldots, y_k\} = f_{\text{verify}}(N, \hat{C}) \).

\subsection{Verification Pipeline}
The verification pipeline consists of two core steps: \textit{candidate expansion} and \textit{contextual revision}. These steps leverage the hierarchical and relational structures of ICD-10-CM to refine and validate model-generated codes.

\subsubsection{Candidate Expansion}
The expansion step broadens the set of candidate codes \( \hat{C} \) by incorporating related codes derived from the tabular tree \( \mathcal{T} \) and the index graph \( \mathcal{G} \). For a candidate \( \hat{c} \), the expanded set is defined as:
\[
\text{Expand}(\hat{c}) = S(\hat{c}) \cup C(\hat{c}) \cup N_1(\hat{c}) \cup N_2(\hat{c}),
\]
where \( S(\hat{c}) \), \( C(\hat{c}) \), \( N_1(\hat{c}) \), and \( N_2(\hat{c}) \) represent the siblings, cousins, 1-step neighbors, and 2-step neighbors of \( \hat{c} \), respectively.

This step ensures that codes related to the initial candidates are explicitly included for further consideration, effectively addressing potential hierarchical misalignments. While complex retrieval methods such as LLM-based semantic search could theoretically be applied, we prioritize structured approaches due to their efficiency, interpretability, and alignment with ICD-10-CM.

In practice, we expand each predicted code by retrieving its siblings \( S(c) \), cousins \( C(c) \), and 1-hop and 2-hop neighbors \( N_1(c) \) and \( N_2(c) \), forming the full candidate set \( \mathcal{C}(c) = S(c) \cup C(c) \cup N_1(c) \cup N_2(c) \). These choices are motivated by the hierarchical and referential structure of ICD-10-CM, and we empirically evaluate different subsets. After de-duplication, expansions remain small relative to the full ICD-10-CM space
($\sim$72k billable codes). In practice, siblings and 1-hop neighbors typically yield
tens of candidates, while cousins and 2-hop neighbors yield a few dozen. This corresponds
to roughly 0.05--0.5\% of the full code list, depending on the seed code and branch.

\subsubsection{Contextual Revision}
The revision step evaluates the expanded candidate set \( \mathcal{C}(\hat{c}) \) by formulating clinical code verification as a multiple-choice task, where the LLM assigns relevance scores to candidate codes based on the clinical note. We investigate several configurations to identify optimal presentation formats, including codes alone, codes paired with descriptions, and descriptions alone. Furthermore, we evaluate the impact of chain-of-thought (CoT) reasoning, in which the model explicitly generates intermediate steps before selecting the final candidate, thus better capturing hierarchical relationships within ICD-10-CM. The final selected code is given by
\(
\hat{c}^{\text{best}} = \arg\max_{c \in \mathcal{C}(\hat{c})} f_{\text{verify}}(N, c),
\)
where \( f_{\text{verify}}(N, c) \) represents the model’s scoring function for a candidate \( c \) given the context \( N \). This verification pipeline can operate as a standalone module or integrate seamlessly into broader medical coding workflows, including end-to-end note-to-code generation systems, auditing procedures, and claim validation processes. It effectively mitigates hierarchical misalignment errors and enhances overall coding accuracy and reliability.

We investigate multiple configurations for the revision process, including presenting the model with only ICD-10-CM codes, codes paired with descriptions, or descriptions alone to assess their impact on verification accuracy. Additionally, we explore the effect of chain-of-thought (CoT) reasoning, where the model generates intermediate steps to better capture hierarchical relationships within ICD-10-CM before selecting the final candidate. These configurations allow us to analyze how contextual information and structured reasoning influence verification performance.

This verification pipeline can function as a standalone module or be integrated into broader medical coding systems. For end-to-end tasks such as clinical note-to-code generation, the verification component complements existing generation methods, reducing errors from hierarchical misalignments and improving overall coding reliability. Similarly, it can be incorporated into auditing workflows or claim validation pipelines to refine outputs and ensure compliance with coding standards. Note that broader expansions increase candidate coverage (i.e., the chance the gold code is included in the candidate set) but also introduce more near-miss distractors, which makes the multiple-choice selection harder.

\section{Experiments}

As mentioned in the introduction, existing datasets such as MIMIC-III/IV are not well-suited for our task. They use ICD-9 codes instead of ICD-10, and their clinical notes are limited to the ICU setting, while codes span the full encounter, making it difficult to verify predictions from a single note. To evaluate the proposed pipeline, we created a new annotated dataset derived from the publicly available ACI-Bench benchmark. We selected outpatient clinical notes and conducted additional expert annotations, assigning gold-standard ICD-10-CM codes and binary verification labels. Each note was double-annotated, with disagreements resolved via arbitration, resulting in 207 annotated examples (67 train, 20 dev, 120 test). Each example includes a clinical note, candidate codes from upstream systems, gold-standard codes, and binary labels indicating validity. Evaluations are performed at the per-note level, aligning with practical requirements in billing, insurance, and clinical decision-making.

\subsection{Models}
We evaluate the pipeline using a diverse set of LLMs for both code generation and verification. These include Claude Haiku and Sonnet (3, 3.5v1, 3.5v2)~\citep{TheC3}, LLaMA (3.1 with 405B, 70B, 8B parameters)~\citep{dubey2024llama}, and Mistral (7B small, large, and Mixtral \(8 \times 7\)B)~\citep{jiang2024mixtral}, covering different architectures, parameter sizes, and training paradigms. We also include PLM-ICD, the conventional SOTA model for ICD-10-CM prediction~\citep{huang-etal-2022-plm}.

\subsection{Evaluation Metrics}

We evaluate the pipeline's performance comprehensively using standard, fuzzy match, and verification metrics. Standard metrics (precision, recall, F1) assess exact matches between predicted and gold-standard codes at the per-note level, aligning with real-world coding practices. However, exact match metrics alone cannot capture hierarchical misalignments within ICD-10-CM codes. Therefore, we introduce fuzzy match metrics leveraging the hierarchical structure of ICD-10-CM. Specifically, we define prefix-$n$ match, which relaxes exact matching by accepting predicted codes whose prefixes match the gold-standard codes up to $n$ steps above the lowest (leaf) level in the ICD hierarchy. For example, prefix-1 match accepts predictions that match at least one step above the leaf node, while prefix-2 allows matching two steps above. Additionally, the prefix overlap ratio measures the weighted hierarchical overlap between predicted and gold-standard codes, considering shared ancestry depth. Finally, we report verification accuracy, directly evaluating the model’s effectiveness at validating pre-assigned candidate codes.

\section{Results}
\label{sec:results}

We evaluate each component of our pipeline individually and assess their combined effect in a full end-to-end setup. As illustrated in Figure~\ref{fig:graph}, our pipeline consists of three main steps: (1) code generation, (2) expansion using the ICD-10-CM hierarchy, and (3) verification to select contextually appropriate codes. The following subsections analyze performance at each step and full pipeline results showing each component's the contribution.

\subsection{Off-the-Shelf LLMs as Medical Coders}

\begin{table}[t]
\centering
\normalsize
\resizebox{\columnwidth}{!}{%
\begin{tabular}{lcccc}
\toprule
\textbf{Generation w/ Simple Prompt} & \textbf{F1 (EM)} & \textbf{F1 (P-1)} & \textbf{F1 (P-2)} & \textbf{POR} \\
\midrule
Haiku-3 & 41.6 & 51.3 & \textbf{54.6} & 62.4 \\
Haiku-3.5 & 40.1 & 48.0 & 52.4 & 62.0 \\
Sonnet-3 & 30.8 & 38.0 & 41.8 & 52.2 \\
Sonnet-3.5v1 & 39.8 & 45.8 & 49.5 & 59.3 \\
Sonnet-3.5v2 & \textbf{43.0} & \textbf{49.6} & 53.8 & \textbf{63.7} \\
LLaMA-3.1-8B & 12.3 & 14.1 & 15.5 & 20.9 \\
LLaMA-3.1-70B & 32.3 & 38.8 & 42.5 & 52.0 \\
LLaMA-3.1-405B & 35.3 & 42.8 & 45.4 & 54.2 \\
Mistral 7B Instruct & 0.94 & 0.99 & 1.4 & 2.11 \\
Mixtral 8x7B Instruct & 26.2 & 33.3 & 38.0 & 45.0 \\
Mistral Small & 26.9 & 32.9 & 38.2 & 46.1 \\
Mistral Large & 35.8 & 42.9 & 47.3 & 55.4 \\
\midrule
PLM-ICD & 24.8 & 35.0 & 38.4 & 50.6 \\
\bottomrule
\end{tabular}%
}
\caption{
Performance of off-the-shelf LLMs on clinical coding with a simple prompt (see~\ref{sec:single_line_prompt}). Exact Match F1 reflects strict correctness, requiring full code matches. Prefix Match F1 scores relax this requirement by allowing matches up to 1 or 2 levels higher in the ICD-10-CM hierarchy, respectively. POR (Prefix Overlap Ratio) quantifies partial correctness based on the shared hierarchical depth between predicted and gold codes.
}

\label{tab:baseline_results}
\end{table}

\vspace{-1mm}

\paragraph{Performance Across Models.} Larger models generally outperformed smaller ones, but improvements were not strictly proportional to scale. As shown in Table~\ref{tab:baseline_results}, Sonnet-3.5v2 achieved the highest performance across all evaluation metrics, surpassing both earlier Sonnet models and larger-scale models like LLaMA-3.1-405B. Among LLaMA models, LLaMA-3.1-405B outperformed the smaller LLaMA-70B, but the gain was smaller compared to the improvement from LLaMA-8B to LLaMA-70B, indicating diminishing returns at extreme scales. Haiku-3 performed competitively, further suggesting that architecture and training paradigms matter significantly alongside model size. Interestingly, Mixtral 7B Instruct had surprisingly poor performance, likely due to mismatch between task formulation and instruction-following capabilities. Overall, these results suggest that scaling alone does not fully resolve complexities of medical coding, especially in outpatient settings characterized by shorter and more varied notes.

\paragraph{Model Generalization and Dataset Transferability.} Models pretrained or fine-tuned on inpatient datasets, such as MIMIC-IV, struggled to generalize effectively to our outpatient dataset. PLM-ICD~\citep{huang-etal-2022-plm}, a state-of-the-art inpatient-focused model, experienced a substantial performance drop in our evaluation. Nevertheless, the Prefix Match F1 and Prefix Overlap Ratio metrics indicated that many of its errors were structurally close to correct codes, highlighting transfer learning challenges inherent in medical coding. Unlike inpatient records, outpatient notes tend to be shorter, fragmented, and contextually ambiguous, making generalization particularly challenging.

\paragraph{Error Analysis: Near Misses and Hallucinations.} Many errors counted as incorrect under Exact Match F1 were near misses, as evidenced by substantial gaps between Exact Match F1 and Prefix-based metrics (e.g., Sonnet-3.5v2 improved from 43.0 Exact Match to 53.8 Prefix-2 Match). These near-miss errors typically occurred within the same hierarchical family, demonstrating that fuzzy match metrics provide a more informative evaluation of clinical coding models. This reinforces the practical need for verification mechanisms to refine outputs and maintain clinical validity, as we will present in Section~\ref{analysis}.

\begin{table}[t]
\centering
\small
\resizebox{\columnwidth}{!}{%
\begin{tabular}{llr}
\toprule
\textbf{Model} & \textbf{Prompt Type} & \textbf{Exact Match F1} \\
\midrule
\multirow{5}{*}{Haiku-3} 
  & Chain-of-Thought & 27.0 \\
  & Detailed Instructions & 36.0 \\
  & Detailed + CoT & 27.2 \\
  & Prompt Decomposition & 40.7 \\
  & Single-Line (Baseline) & \textbf{41.6} \\
\midrule
\multirow{5}{*}{Sonnet-3.5v1} 
  & Chain-of-Thought & 47.4 \\
  & Detailed Instructions & 47.9 \\
  & Detailed + CoT & \textbf{55.6} \\
  & Prompt Decomposition & 42.3 \\
  & Single-Line (Baseline) & 39.8 \\
\bottomrule
\end{tabular}%
}
\caption{
Impact of prompt engineering on Haiku-3 and Sonnet-3.5v1. Prompt types include structured reasoning (CoT), detailed instructions, prompt decomposition, and a minimal single-line baseline in~\autoref{prompt_eng}. All prompts except the baseline ask models to predict both the code and description with priority given to code.
}
\label{tab:prompt_results}
\end{table}



\subsection{Lightweight Methods to Improve LLMs}

\paragraph{Prompt Engineering Benefits Stronger Models.} Prompt engineering significantly improved performance for stronger models like Sonnet-3.5v1, with gains exceeding 15 points in Exact Match F1 (from 39.8 to 55.6). As shown in Table~\ref{tab:prompt_results} (more in Appendix~\ref{app:addtional_results:prompt}), the best performance was achieved by combining detailed instructions with chain-of-thought (CoT) reasoning. In contrast, Haiku-3 showed minimal improvement, with the simplest single-line baseline outperforming all other prompt types. This suggests that prompt engineering is more effective when models have sufficient capacity to follow and benefit from structured reasoning.

\paragraph{Fine-Tuning Enables Robust Adaptation.} Fine-tuning Haiku-3 on a small, high-quality dataset (67 examples) led to substantial performance gains. As shown in Table~\ref{tab:fine_tune_results}, the best results were achieved when the model was trained to generate both codes and their corresponding descriptions, with code appearing first (from 40.6 to 56.9). However, fine-tuning was highly sensitive to the output format: training on code alone resulted in a collapse in performance (from 32.7 to 0.0). These results highlight that fine-tuning can be highly effective but requires careful prompt-target design.

\begin{table}[t]
\centering
\small
\resizebox{\columnwidth}{!}{%
\begin{tabular}{llr}
\toprule
\textbf{Prompt Setting} & \textbf{Model} & \textbf{Exact Match F1} \\
\midrule
\multirow{2}{*}{Code Only} 
  & Haiku-3 & 32.7 \\
  & + Fine-tuned & ($\downarrow$32.7) \space 0.0  \\
\midrule
\multirow{2}{*}{Description Only} 
  & Haiku-3 & 10.3 \\
  & + Fine-tuned & ($\downarrow$0.7) 11.0 \\
\midrule
\multirow{2}{*}{Description → Code} 
  & Haiku-3 & 30.4 \\
  & + Fine-tuned & ($\uparrow$21.6) 52.0 \\
\midrule
\multirow{2}{*}{Code → Description} 
  & Haiku-3 & 40.6 \\
  & + Fine-tuned & ($\uparrow$16.3) \textbf{56.9} \\
\bottomrule
\end{tabular}%
}
\caption{
Fine-tuning results for Haiku-3 across different generation targets. Configurations vary in whether the model generates codes, descriptions, or both (in different orders). Inference uses the same format as the corresponding prompt. Scores in parentheses denote the performance changes (delta) after fine-tuning.
}

\label{tab:fine_tune_results}
\end{table}


\subsection{Verification and Full Pipeline Evaluation}

\begin{table}[t]
\centering
\small
\resizebox{\columnwidth}{!}{%
\begin{tabular}{llr}
\toprule
\textbf{Expansion Type} & \textbf{Prompt Variant} & \textbf{Accuracy (\%)} \\
\midrule
\multirow{4}{*}{Siblings \(S(c)\)} 
  & Code + Description  & 82.1 \\
  & Code Only & 82.5 \\
  & Chain-of-Thought (CoT) & 82.2 \\
  & Description Only & \textbf{88.3} \\
\midrule
\multirow{4}{*}{Cousins \(C(c)\)} 
  & Code + Description  & 87.4 \\
  & Code Only & 86.6 \\
  & Chain-of-Thought (CoT) & 88.5 \\
  & Description Only & \textbf{90.3} \\
\midrule
\multirow{4}{*}{1-Hop Neighbors \(N_1(c)\)} 
  & Code + Description  & 82.6 \\
  & Code Only & 83.4 \\
  & Chain-of-Thought (CoT) & 82.3 \\
  & Description Only & \textbf{85.8} \\
\midrule
\multirow{4}{*}{2-Hop Neighbors \(N_2(c)\)} 
  & Code + Description  & 85.4 \\
  & Code Only & 85.4 \\
  & Chain-of-Thought (CoT) & 85.3 \\
  & Description Only & \textbf{87.4} \\
\midrule
\multirow{4}{*}{All Combined} 
  & Code + Description  & 77.3 \\
  & Code Only & 78.9 \\
  & Chain-of-Thought (CoT) & 78.9 \\
  & Description Only & \textbf{82.2} \\
\bottomrule
\end{tabular}%
}
\caption{
Verification accuracy (\%) of Sonnet-3.5v2 under different candidate expansion types and prompt formats. Expansions include siblings \(S(c)\), cousins \(C(c)\), 1-hop \(N_1(c)\), and 2-hop \(N_2(c)\) neighbors as defined in Section~\ref{structure}. All use multiple-choice formatting; prompt variants vary in how candidate codes are presented.
}
\label{tab:verification_results}
\end{table}



\paragraph{Descriptions Boost Verification Accuracy.} 
In the standalone verification task, the model receives a clinical note and a set of candidate codes expanded from a gold code, ensuring the correct answer is always present, with 100\% as the upper bound. As shown in Table~\ref{tab:verification_results}, presenting only descriptions consistently yields the highest accuracy across all expansions. For instance, with cousin expansions \(C(c)\), Sonnet-3.5v2 achieves 90.3\% accuracy using description-only input. Chain-of-thought (CoT) reasoning, by contrast, fails to improve performance and can slightly reduce it. This suggests lexical context is more useful than structured reasoning in this setting. Verification accuracy also varies by expansion. Candidates from sibling sets \(S(c)\) are harder to distinguish due to semantic similarity, while 2-hop neighbors \(N_2(c)\) are more diverse and easier to filter. We observe a trade-off that broader expansions can raise accuracy but increase candidate count and inference cost. Combining all expansion types increases coverage but also enlarges the candidate set with near-miss distractors, which lowers this conditional accuracy. In contrast, the full pipeline results in Table~\ref{tab:full_pipeline_results} optimize coverage and selection jointly, yielding consistent gains in end-to-end F1.

\begin{table}[t]
\centering
\resizebox{\columnwidth}{!}{%
\begin{tabular}{lrr|r}
\toprule
\textbf{Model} & \textbf{Generation} & \textbf{+ Verification} & \textbf{+ Oracle} \\
\midrule
Haiku-3 & 41.6 & \textbf{47.2} & 54.1 \\
Haiku-3 (Fine-tuned) & 56.9 & \textbf{57.6} & 67.3 \\
Sonnet-3.5v1 & \textbf{55.6} & 55.5 & 66.4 \\
PLM-ICD & 24.8 & \textbf{29.4} & 31.9 \\
\bottomrule
\end{tabular}
}
\caption{
Performance of the full clinical coding pipeline. The “+ Verification” setting uses expanded candidates (\(S(c)\), \(C(c)\), \(N_1(c)\), \(N_2(c)\)) with LLM-based Description Only verification in~\autoref{tab:verification_results}. “+ Oracle ” replaces verification to select only the ground-truth labels from the expansion (if in candidates) as the upper-bound.
}
\label{tab:full_pipeline_results}
\end{table}


\paragraph{Full Pipeline Improves End-to-End Accuracy.}
Generated codes are expanded using all structural relationships (\(S(c), C(c), N_1(c), N_2(c)\)) before verification. As shown in Table~\ref{tab:full_pipeline_results}\footnote{Oracle performance is below 100\% because expansions are applied to
model-predicted seeds. If the gold code lies outside the expanded branch,
it cannot be recovered even under oracle selection.}
, this improves performance across all models. Haiku-3 rises from 41.6 to 47.2 F1, PLM-ICD from 24.8 to 29.4, and fine-tuned Haiku-3 reaches 57.6 F1, an increase of 16 points over the baseline. Stronger models like Sonnet-3.5v1 also benefit or maintain performance. Oracle selection, which uses gold labels to pick the correct candidate, shows an upper bound of up to 67.3 F1. These results confirm that verification is a lightweight, model-agnostic component that reliably improves clinical coding.

\subsection{Insights from a Clinical Perspective}
\label{analysis}

Clinical coding follows strict domain-specific rules that LLMs are not trained to model explicitly. Several consistent patterns emerged: First, models frequently assign codes for signs and symptoms (e.g., ICD-10-CM “R” codes) even when the underlying etiology is already provided. For example, predicting \texttt{R26.2} (difficulty walking) is incorrect when \texttt{M51.27} (disc displacement) is also present and fully explains the symptom. Second, models often code for possible or suspected conditions, which is discouraged in outpatient settings where only confirmed diagnoses should be coded. More broadly, both models tend to overcode, generating plausible but unnecessary entries. While this is preferable to undercoding (as human coders can remove extras more easily than spotting omissions), it can introduce noise and mislead downstream systems. We also observed clinically meaningful hallucinations. In one case, a hemoglobin lab result was misread as a hemoglobin A1c, leading the model to falsely assign a diabetes code. While it is unclear how common this is, such errors reveal a tendency to overgeneralize from surface cues.

\paragraph{Verification Enhances Clinical Reliability.} Our full pipeline mitigates these issues in practice. In the example shown in Figure~\ref{fig:graph}, the fine-tuned model incorrectly predicted \texttt{M25.561} (Pain in right knee) for a case describing left knee pain. Verification corrected this, identifying \texttt{M25.562} (Pain in left knee) as a sibling and selecting the appropriate code. In another example, the model predicted \texttt{R78.71} (Abnormal lead level in blood) when the correct code was \texttt{R79.9} (Abnormal finding of blood chemistry, unspecified). Verification revised it to \texttt{R78.9} (Finding of unspecified substance, not normally found in blood), not exact, but semantically closer. These examples highlight the value of structured verification and expert-guided evaluation. Quantitative metrics provide signal, but qualitative clinical insight remains essential for safe and trustworthy medical AI in the healthcare domain.

Overall, our error analysis shows that most near-miss mistakes occur late in the ICD-10-CM
hierarchy, typically at the leaf or leaf--1 levels. Common cases include laterality
(e.g., left vs.\ right), presence vs.\ absence of complications, or similar fine-grained
distinctions. This pattern is consistent with our prefix-$n$ metrics, which capture
hierarchical closeness of predictions.

\section{Related Work}

Clinical coding has traditionally relied on specialized models such as ClinicalBERT \citep{huang2019clinicalbert} and PLM-ICD \citep{huang-etal-2022-plm}. Recent studies using off-the-shelf LLMs report poor performance under exact match metrics \citep{boyle2023automated, soroush2024large}, but do not address hierarchical misalignment, which is the focus of this work. Approaches like LLM-guided tree search \citep{Klang2024assessing} and retrieval-augmented generation \citep{baksi2024medcoder} improve accuracy but are computationally expensive and hard to maintain as coding standards evolve. \citet{baksi2024medcoder} also release a dataset, but their annotations are locally scoped and less consistent; third-party clinical experts rated our double expert annotated dataset as higher quality. We view their work as complementary. Lightweight adaptation methods such as prompt engineering and small-scale fine-tuning have shown promise for domain-specific tasks \citep{brown2020language}, and we explore them for clinical coding. Structured verification is common in fact-checking \citep{yuan-vlachos-2024-zero} but underexplored in this domain. Existing pipelines include verification implicitly \citep{yang2023surpassing}, but it has not been evaluated as a standalone component. Finally, most evaluations rely on exact match and overlook the hierarchical nature of ICD-10.

\section{Conclusion}
We presented lightweight methods to improve the reliability of LLM-based clinical coding. Our experiments showed that prompt engineering and small-scale fine-tuning can yield substantial gains without the computational overhead of large-scale retrieval or search. To further address common near-miss errors, we introduced \emph{clinical code verification}, a model-agnostic pipeline that leverages the ICD-10-CM hierarchy to refine predictions and correct hierarchical misalignments. A key contribution of this work is the release of a new expert double-annotated dataset 
of outpatient clinical notes with ICD-10-CM codes. This resource provides high-quality supervision in a setting underrepresented in existing benchmarks, enabling more rigorous and clinically meaningful evaluation.

From a clinical perspective, verification mitigates frequent mistakes such as miscoding of laterality or symptom versus etiology, offering a practical safeguard against errors with financial and clinical implications. Our results highlight that coding reliability depends not only on raw generation accuracy but also on structured mechanisms for refinement and evaluation.

\newpage

\section*{Limitations}

While our approach demonstrates meaningful improvements in clinical coding, several limitations should be acknowledged. First, our experiments are conducted on a small, expert-annotated dataset of outpatient clinical notes, which may not fully capture the diversity of coding scenarios across inpatient settings or other healthcare domains. Second, the verification step operates on candidate codes produced by LLMs and is thus influenced by the quality and biases of the underlying generation model. Third, the fine-tuning experiments are performed on a limited dataset of 67 examples. While this enables controlled evaluation under expert supervision, it may limit the generalizability of the observed improvements. Finally, although our results are promising, they do not imply readiness for clinical deployment. Additional validation with domain experts and assessments of robustness and safety would be necessary for real-world use. All prompts, 207-note dataset, ICD expansion scripts, and evaluation code will be released under MIT licence upon notification and approval.

\section*{Ethical Considerations}
The use of LLMs in clinical coding raises important ethical concerns, particularly regarding patient privacy, bias, accountability, and the potential for harm due to incorrect coding. While automated coding can enhance efficiency, it also introduces risks if errors propagate through billing, insurance claims, and clinical decision-making.

\paragraph{Patient Privacy and Data Security.} Clinical coding involves sensitive patient information, making privacy and data security paramount. Although our approach does not require direct patient identifiers, LLMs trained on medical text may still encode latent biases or inadvertently reveal sensitive details. Ensuring compliance with healthcare regulations such as HIPAA (U.S.) and GDPR (EU) is critical for responsible deployment.

\paragraph{Bias and Fairness.} LLMs inherit biases from their training data, which may lead to disparities in coding accuracy across different demographic groups, clinical conditions, or healthcare settings. If left unchecked, these biases can disproportionately affect underrepresented populations, leading to systematic errors in billing and treatment documentation. Our verification step aims to reduce coding errors, but future work should include bias audits and fairness-aware training approaches to mitigate these risks.

\paragraph{Accountability and Human Oversight.} Clinical coding impacts reimbursement, resource allocation, and patient care. Over-reliance on AI without human oversight could introduce systemic errors, particularly if models reinforce preexisting coding patterns without adapting to evolving medical guidelines. Our verification approach is designed as an assistive tool, not a replacement for human coders. We advocate for AI-assisted workflows where models support human decision-making rather than fully automate coding.

\paragraph{Clinical Validation and Deployment Risks.} Deploying AI-assisted coding systems in healthcare settings requires rigorous validation. Misclassifications in ICD-10-CM codes can have financial consequences, such as incorrect billing. Future studies should involve clinical experts in evaluating model predictions, ensuring transparency in decision-making, and establishing safety protocols before real-world adoption.

\paragraph{Long-Term Implications.} As LLMs continue to evolve, ensuring ethical AI deployment in healthcare will require continuous monitoring, interpretability improvements, and alignment with medical best practices. Stakeholders, including healthcare providers, regulators, and AI developers, must collaborate to establish guidelines for responsible AI integration in clinical workflows.

\bibliography{custom}

\begin{thebibliography}{22}
\providecommand{\natexlab}[1]{#1}

\bibitem[{Adams et~al.(2022)Adams, Shing, Sun, Winestock, McKeown, and Elhadad}]{adams-etal-2022-learning}
Griffin Adams, Han-Chin Shing, Qing Sun, Christopher Winestock, Kathleen McKeown, and No{\'e}mie Elhadad. 2022.
\newblock \href {https://doi.org/10.18653/v1/2022.findings-emnlp.296} {Learning to revise references for faithful summarization}.
\newblock In \emph{Findings of the Association for Computational Linguistics: EMNLP 2022}, pages 4009--4027, Abu Dhabi, United Arab Emirates. Association for Computational Linguistics.

\bibitem[{Anthropic(2024)}]{TheC3}
Anthropic. 2024.
\newblock \href {https://api.semanticscholar.org/CorpusID:268232499} {The claude 3 model family: Opus, sonnet, haiku}.

\bibitem[{Baksi et~al.(2024)Baksi, Soba, Higgins, Saini, Wood, Cook, Scott, Pudota, Weninger, Bowen et~al.}]{baksi2024medcoder}
Krishanu~Das Baksi, Elijah Soba, John~J Higgins, Ravi Saini, Jaden Wood, Jane Cook, Jack Scott, Nirmala Pudota, Tim Weninger, Edward Bowen, et~al. 2024.
\newblock Medcoder: A generative ai assistant for medical coding.
\newblock \emph{arXiv preprint arXiv:2409.15368}.

\bibitem[{Boyle et~al.(2023)Boyle, Kascenas, Lok, Liakata, and O'Neil}]{boyle2023automated}
Joseph~S Boyle, Antanas Kascenas, Pat Lok, Maria Liakata, and Alison~Q O'Neil. 2023.
\newblock Automated clinical coding using off-the-shelf large language models.
\newblock \emph{arXiv preprint arXiv:2310.06552}.

\bibitem[{Brown et~al.(2020)Brown, Mann, Ryder, Subbiah, Kaplan, Dhariwal, Neelakantan, Shyam, Sastry, Askell et~al.}]{brown2020language}
Tom Brown, Benjamin Mann, Nick Ryder, Melanie Subbiah, Jared~D Kaplan, Prafulla Dhariwal, Arvind Neelakantan, Pranav Shyam, Girish Sastry, Amanda Askell, et~al. 2020.
\newblock Language models are few-shot learners.
\newblock \emph{Advances in neural information processing systems}, 33:1877--1901.

\bibitem[{DeYoung et~al.(2022)DeYoung, Shing, Kong, Winestock, and Shivade}]{deyoung2022entity}
Jay DeYoung, Han-Chin Shing, Luyang Kong, Christopher Winestock, and Chaitanya Shivade. 2022.
\newblock Entity anchored icd coding.
\newblock \emph{arXiv preprint arXiv:2208.07444}.

\bibitem[{Dubey et~al.(2024)Dubey, Jauhri, Pandey, Kadian, Al-Dahle, Letman, Mathur, Schelten, Yang, Fan et~al.}]{dubey2024llama}
Abhimanyu Dubey, Abhinav Jauhri, Abhinav Pandey, Abhishek Kadian, Ahmad Al-Dahle, Aiesha Letman, Akhil Mathur, Alan Schelten, Amy Yang, Angela Fan, et~al. 2024.
\newblock The llama 3 herd of models.
\newblock \emph{arXiv preprint arXiv:2407.21783}.

\bibitem[{Hirsch et~al.(2016)Hirsch, Nicola, McGinty, Liu, Barr, Chittle, and Manchikanti}]{hirsch2016icd}
JA~Hirsch, G~Nicola, G~McGinty, RW~Liu, RM~Barr, MD~Chittle, and L~Manchikanti. 2016.
\newblock Icd-10: history and context.
\newblock \emph{American Journal of Neuroradiology}, 37(4):596--599.

\bibitem[{Huang et~al.(2022)Huang, Tsai, and Chen}]{huang-etal-2022-plm}
Chao-Wei Huang, Shang-Chi Tsai, and Yun-Nung Chen. 2022.
\newblock \href {https://doi.org/10.18653/v1/2022.clinicalnlp-1.2} {{PLM}-{ICD}: Automatic {ICD} coding with pretrained language models}.
\newblock In \emph{Proceedings of the 4th Clinical Natural Language Processing Workshop}, pages 10--20, Seattle, WA. Association for Computational Linguistics.

\bibitem[{Huang et~al.(2019)Huang, Altosaar, and Ranganath}]{huang2019clinicalbert}
Kexin Huang, Jaan Altosaar, and Rajesh Ranganath. 2019.
\newblock Clinicalbert: Modeling clinical notes and predicting hospital readmission.
\newblock \emph{arXiv:1904.05342}.

\bibitem[{Jha et~al.(2009)Jha, DesRoches, Campbell, Donelan, Rao, Ferris, Shields, Rosenbaum, and Blumenthal}]{jha2009use}
Ashish~K Jha, Catherine~M DesRoches, Eric~G Campbell, Karen Donelan, Sowmya~R Rao, Timothy~G Ferris, Alexandra Shields, Sara Rosenbaum, and David Blumenthal. 2009.
\newblock Use of electronic health records in us hospitals.
\newblock \emph{New England Journal of Medicine}, 360(16):1628--1638.

\bibitem[{Jiang et~al.(2024)Jiang, Sablayrolles, Roux, Mensch, Savary, Bamford, Chaplot, Casas, Hanna, Bressand et~al.}]{jiang2024mixtral}
Albert~Q Jiang, Alexandre Sablayrolles, Antoine Roux, Arthur Mensch, Blanche Savary, Chris Bamford, Devendra~Singh Chaplot, Diego de~las Casas, Emma~Bou Hanna, Florian Bressand, et~al. 2024.
\newblock Mixtral of experts.
\newblock \emph{arXiv preprint arXiv:2401.04088}.

\bibitem[{Johnson et~al.(2016)Johnson, Pollard, Shen, Lehman, Feng, Ghassemi, Moody, Szolovits, Anthony~Celi, and Mark}]{johnson2016mimic}
Alistair~EW Johnson, Tom~J Pollard, Lu~Shen, Li-wei~H Lehman, Mengling Feng, Mohammad Ghassemi, Benjamin Moody, Peter Szolovits, Leo Anthony~Celi, and Roger~G Mark. 2016.
\newblock Mimic-iii, a freely accessible critical care database.
\newblock \emph{Scientific data}, 3(1):1--9.

\bibitem[{Klang et~al.(2024)Klang, Tessler, Apakama, Abbott, Glicksberg, Arnold, Moses, Sakhuja, Soroush, Charney, Reich, McGreevy, Gavin, Carr, Freeman, and Nadkarni}]{Klang2024assessing}
Eyal Klang, Idit Tessler, Donald~U Apakama, Ethan Abbott, Benjamin~S Glicksberg, Monique Arnold, Akini Moses, Ankit Sakhuja, Ali Soroush, Alexander~W Charney, David~L. Reich, Jolion McGreevy, Nicholas Gavin, Brendan Carr, Robert Freeman, and Girish~N Nadkarni. 2024.
\newblock \href {https://doi.org/10.1101/2024.10.15.24315526} {Assessing retrieval-augmented large language model performance in emergency department icd-10-cm coding compared to human coders}.
\newblock \emph{medRxiv}.

\bibitem[{Liu et~al.(2022)Liu, Perez-Concha, Nguyen, Bennett, and Jorm}]{liu2022hierarchical}
Leibo Liu, Oscar Perez-Concha, Anthony Nguyen, Vicki Bennett, and Louisa Jorm. 2022.
\newblock Hierarchical label-wise attention transformer model for explainable icd coding.
\newblock \emph{Journal of biomedical informatics}, 133:104161.

\bibitem[{Shickel et~al.(2017)Shickel, Tighe, Bihorac, and Rashidi}]{shickel2017deep}
Benjamin Shickel, Patrick~James Tighe, Azra Bihorac, and Parisa Rashidi. 2017.
\newblock Deep ehr: a survey of recent advances in deep learning techniques for electronic health record (ehr) analysis.
\newblock \emph{IEEE journal of biomedical and health informatics}, 22(5):1589--1604.

\bibitem[{Shing et~al.(2021)Shing, Shivade, Pourdamghani, Nan, Resnik, Oard, and Bhatia}]{shing2021clinicalencountersummarizationlearning}
Han-Chin Shing, Chaitanya Shivade, Nima Pourdamghani, Feng Nan, Philip Resnik, Douglas Oard, and Parminder Bhatia. 2021.
\newblock \href {https://arxiv.org/abs/2104.13498} {Towards clinical encounter summarization: Learning to compose discharge summaries from prior notes}.
\newblock \emph{Preprint}, arXiv:2104.13498.

\bibitem[{Soroush et~al.(2024)Soroush, Glicksberg, Zimlichman, Barash, Freeman, Charney, Nadkarni, and Klang}]{soroush2024large}
Ali Soroush, Benjamin~S Glicksberg, Eyal Zimlichman, Yiftach Barash, Robert Freeman, Alexander~W Charney, Girish~N Nadkarni, and Eyal Klang. 2024.
\newblock Large language models are poor medical coders—benchmarking of medical code querying.
\newblock \emph{NEJM AI}, 1(5):AIdbp2300040.

\bibitem[{Yang et~al.(2023)Yang, Batra, Stremmel, and Halperin}]{yang2023surpassing}
Zhichao Yang, Sanjit~Singh Batra, Joel Stremmel, and Eran Halperin. 2023.
\newblock Surpassing gpt-4 medical coding with a two-stage approach.
\newblock \emph{arXiv preprint arXiv:2311.13735}.

\bibitem[{Yih et~al.(2023)Yih, Daley, Duffy, Fireman, McClure, Nelson, Qian, Smith, Vazquez-Benitez, Weintraub et~al.}]{yih2023broad}
W~Katherine Yih, Matthew~F Daley, Jonathan Duffy, Bruce Fireman, David McClure, Jennifer Nelson, Lei Qian, Ning Smith, Gabriela Vazquez-Benitez, Eric Weintraub, et~al. 2023.
\newblock A broad assessment of covid-19 vaccine safety using tree-based data-mining in the vaccine safety datalink.
\newblock \emph{Vaccine}, 41(3):826--835.

\bibitem[{Yim et~al.(2023)Yim, Fu, Ben~Abacha, Snider, Lin, and Yetisgen}]{yim2023aci}
Wen-wai Yim, Yujuan Fu, Asma Ben~Abacha, Neal Snider, Thomas Lin, and Meliha Yetisgen. 2023.
\newblock Aci-bench: a novel ambient clinical intelligence dataset for benchmarking automatic visit note generation.
\newblock \emph{Scientific Data}, 10(1):586.

\bibitem[{Yuan and Vlachos(2024)}]{yuan-vlachos-2024-zero}
Moy Yuan and Andreas Vlachos. 2024.
\newblock \href {https://doi.org/10.18653/v1/2024.kallm-1.11} {Zero-shot fact-checking with semantic triples and knowledge graphs}.
\newblock In \emph{Proceedings of the 1st Workshop on Knowledge Graphs and Large Language Models (KaLLM 2024)}, pages 105--115, Bangkok, Thailand. Association for Computational Linguistics.

\end{thebibliography}

\newpage
\appendix

\section{ICD-10-CM Hierarchical Structure}
\label{appendix:icd_structure}

ICD-10-CM codes follow a hierarchical structure defined by the official tabular list. Each node in the tree represents a diagnostic category, and only leaf nodes are billable. This structure introduces challenges for automatic coding, as models must distinguish between closely related codes that differ in specificity and billing eligibility.

\begin{figure}[h]
    \centering
    \includegraphics[width=\columnwidth]{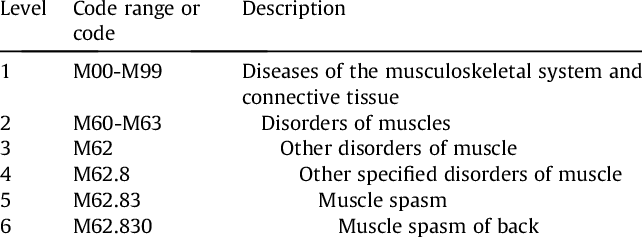}
    \caption{
    Example fragment of the ICD-10-CM hierarchy, adapted from \citet{yih2023broad}. Only leaf nodes are billable. Nodes with the same parent are considered siblings; those with the same grandparent are cousins.
    }
    \label{fig:icd_tree}
\end{figure}

Figure~\ref{fig:icd_tree} illustrates a simplified example of the ICD-10-CM tree structure, adapted from \citet{yih2023broad}. Nodes are connected via parent-child relationships, with each level corresponding to a finer-grained diagnostic category. Siblings (\(S(c)\)) share the same parent, and cousins (\(C(c)\)) share a grandparent. Understanding these structural relations is crucial for evaluating both exact and near-miss predictions in clinical coding.

\section{Additional Results}
\label{app:addtional_results}

This Appendix provides extended results for verification, prompt engineering, fine-tuning, and the full pipeline evaluation. These results complement the main text by presenting a detailed breakdown of model performance across different configurations.

\subsection{Prompt Engineering Results Across All Models}
\label{app:addtional_results:prompt}
Table~\ref{tab:appendix_prompt_results_full} presents the complete results of our prompt engineering experiments, which examined how different prompt engineering techniques affect model performance. Our findings indicate that prompt effectiveness varies across different models, and there is no single prompt format that works best for all models.


\begin{table*}
\resizebox{2\columnwidth}{!}{%
\begin{tabular}{llllll}
\toprule
 &  & Exact Match F1 & Prefix Match F1 (Prefix-1) & Prefix Match F1 (Prefix-2) & Prefix Overlap Ratio \\
Model & Prompt Type &  &  &  &  \\
\midrule
\multirow[t]{5}{*}{Haiku-3} & Chain-of-Thought & 27.000 & 36.588 & 39.134 & 46.953 \\
 & Detailed + CoT & 27.161 & 32.321 & 34.839 & 41.760 \\
 & Detailed Instructions & 36.048 & 42.941 & 45.691 & 53.159 \\
 & Prompt Decomposition & 40.720 & 46.969 & 49.890 & 57.692 \\
 & Single-Line (Baseline) & \textbf{41.639} & \textbf{51.294} & \textbf{54.640} & \textbf{62.407} \\
\midrule
\multirow[t]{5}{*}{Haiku-3.5} & Chain-of-Thought & 36.521 & 44.361 & 48.030 & 60.006 \\
 & Detailed + CoT & \textbf{41.438} & \textbf{49.553} & \textbf{56.421} & 60.541 \\
 & Detailed Instructions & 39.338 & 47.169 & 50.203 & 57.918 \\
 & Prompt Decomposition & 14.194 & 18.277 & 21.055 & 21.631 \\
 & Single-Line (Baseline) & 40.115 & 47.976 & 52.418 & \textbf{61.996} \\
\midrule
\multirow[t]{5}{*}{Sonnet-3} & Chain-of-Thought & 34.034 & 40.388 & 43.363 & 52.709 \\
 & Detailed + CoT & \textbf{41.465} & \textbf{49.902} & \textbf{52.069} & 60.511 \\
 & Detailed Instructions & 40.294 & 48.202 & 51.758 & \textbf{61.143} \\
 & Prompt Decomposition & 36.158 & 44.007 & 48.176 & 56.936 \\
 & Single-Line (Baseline) & 30.819 & 38.040 & 41.831 & 52.232 \\
\midrule
\multirow[t]{5}{*}{Sonnet-3.5v1} & Chain-of-Thought & 47.425 & 55.573 & 59.313 & 70.617 \\
 & Detailed + CoT & \textbf{55.643} & \textbf{62.429} & \textbf{65.095} & \textbf{74.463} \\
 & Detailed Instructions & 47.868 & 56.401 & 60.687 & 70.608 \\
 & Prompt Decomposition & 42.323 & 47.788 & 51.564 & 61.854 \\
 & Single-Line (Baseline) & 39.834 & 45.805 & 49.478 & 59.260 \\
\midrule
\multirow[t]{5}{*}{Sonnet-3.5v2} & Chain-of-Thought & 40.458 & 48.892 & 55.799 & 66.009 \\
 & Detailed + CoT & \textbf{50.587} & \textbf{60.594} & \textbf{63.668} & \textbf{74.632} \\
 & Detailed Instructions & 45.744 & 53.348 & 57.824 & 68.607 \\
 & Prompt Decomposition & 42.618 & 48.240 & 53.067 & 62.258 \\
 & Single-Line (Baseline) & 42.982 & 49.581 & 53.751 & 63.662 \\
\midrule
\multirow[t]{5}{*}{LLaMA-3.1-8B} & Chain-of-Thought & 15.675 & 18.552 & 20.929 & 27.428 \\
 & Detailed + CoT & 21.826 & 27.789 & 29.087 & 36.424 \\
 & Detailed Instructions & 23.057 & 27.748 & 31.301 & 40.424 \\
 & Prompt Decomposition & \textbf{24.404} & \textbf{30.789} & \textbf{35.606} & \textbf{46.259} \\
 & Single-Line (Baseline) & 12.261 & 14.118 & 15.481 & 20.898 \\
\midrule
\multirow[t]{5}{*}{LLaMA-3.1-70B} & Chain-of-Thought & 32.491 & 40.721 & 45.238 & 55.714 \\
 & Detailed + CoT & \textbf{36.121} & \textbf{44.992} & \textbf{49.367} & \textbf{59.396} \\
 & Detailed Instructions & 31.641 & 38.258 & 42.599 & 54.027 \\
 & Prompt Decomposition & 32.235 & 38.052 & 42.056 & 50.524 \\
 & Single-Line (Baseline) & 32.267 & 38.836 & 42.453 & 51.978 \\
\midrule
\multirow[t]{5}{*}{LLaMA-3.1-405b} & Chain-of-Thought & 33.251 & 40.289 & 43.554 & 52.090 \\
 & Detailed + CoT & 34.036 & 42.810 & 45.014 & 52.819 \\
 & Detailed Instructions & 34.406 & 39.832 & \textbf{45.527} & 53.264 \\
 & Prompt Decomposition & 34.158 & 41.176 & 44.986 & 52.723 \\
 & Single-Line (Baseline) & \textbf{35.347} & \textbf{42.826} & 45.444 & \textbf{54.194} \\
\midrule
\multirow[t]{5}{*}{Mistral 7B Instruct} & Chain-of-Thought & 17.669 & 22.890 & 25.381 & 32.931 \\
 & Detailed + CoT & 15.279 & 21.091 & 22.492 & 28.814 \\
 & Detailed Instructions & 12.882 & 17.735 & 19.263 & 27.121 \\
 & Prompt Decomposition & \textbf{19.960} & \textbf{24.881} & \textbf{27.057} & \textbf{35.601} \\
 & Single-Line (Baseline) & 0.941 & 0.994 & 1.405 & 2.114 \\
\midrule
\multirow[t]{5}{*}{Mixtral 8x7B Instruct} & Chain-of-Thought & 26.417 & 35.803 & 38.382 & 45.245 \\
 & Detailed + CoT & 29.167 & 38.784 & 42.711 & 50.884 \\
 & Detailed Instructions & 29.114 & \textbf{39.390} & 45.617 & 53.742 \\
 & Prompt Decomposition & \textbf{33.222} & 39.097 & \textbf{46.115} & \textbf{53.795} \\
 & Single-Line (Baseline) & 26.198 & 33.291 & 37.983 & 44.990 \\
\midrule
\multirow[t]{5}{*}{Mistral Small} & Chain-of-Thought & 22.146 & 29.621 & 32.845 & 37.718 \\
 & Detailed + CoT & 29.092 & 35.059 & 38.732 & 45.652 \\
 & Detailed Instructions & 19.323 & 24.688 & 27.717 & 35.906 \\
 & Prompt Decomposition & \textbf{29.775} & \textbf{37.511} & \textbf{43.035} & \textbf{51.553} \\
 & Single-Line (Baseline) & 26.863 & 32.878 & 38.250 & 46.128 \\
\midrule
\multirow[t]{5}{*}{Mistral Large} & Chain-of-Thought & 34.332 & 44.899 & 50.683 & 57.119 \\
 & Detailed + CoT & \textbf{36.641} & \textbf{47.293} & \textbf{53.569} & \textbf{60.415} \\
 & Detailed Instructions & 34.264 & 43.181 & 48.304 & 55.545 \\
 & Prompt Decomposition & 36.328 & 47.204 & 51.553 & 57.968 \\
 & Single-Line (Baseline) & 35.754 & 42.880 & 47.286 & 55.443 \\
\midrule
\bottomrule
\end{tabular}
}
\caption{Performance of off-the-shelf LLMs on clinical coding. Exact Match F1 reflects strict correctness, while Prefix Match F1 evaluates correctness at different hierarchical levels (1 or 2 levels). Prefix Overlap Ratio measures partial correctness weighted by shared hierarchical depth.}
\label{tab:appendix_prompt_results_full}
\end{table*}

\subsection{Verification Results Across All Models}
\label{app:addtional_results:verification}
Table~\ref{tab:appendix_verification_results} presents verification accuracy for all tested models under two prompt configurations: one using Code + Description and another adding a Chain-of-Thought (CoT) step. These results demonstrate the robustness of our verification approach across different model architectures and parameter scales. While some models (e.g., Sonnet-3 and Mistral Small) show clear benefits from Chain-of-Thought reasoning, we do not observe a consistent pattern across different models.

\begin{table*}[ht]
\centering
\small
\resizebox{2\columnwidth}{!}{%
\begin{tabular}{llrrrrr}
\toprule
 & Expansion Type & Siblings \(S(c)\) & Cousins \(C(c)\) & 1-Hop Neighbors \(N_1(c)\) & 2-Hop Neighbors \(N_2(c)\) & All Combined \\
Model & Prompt Variant &  &  &  &  &  \\
\midrule
\multirow[t]{2}{*}{Haiku-3} & Code + Description & 52.2 & 51.4 & 59.5 & 50.2 & 41.3 \\
 & + Chain-of-Thought (CoT)  & 58.4 & 56.9 & 56.9 & 50.0 & 40.1 \\
\midrule
\multirow[t]{2}{*}{Haiku-3.5} & Code + Description & 74.9 & 80.7 & 80.6 & 82.4 & 70.2 \\
 & + Chain-of-Thought (CoT)  & 78.9 & 77.9 & 81.7 & 76.7 & 68.4 \\
\midrule
\multirow[t]{2}{*}{Sonnet-3} & Code + Description & 66.0 & 64.2 & 68.4 & 54.9 & 43.9 \\
 & + Chain-of-Thought (CoT)  & 74.4 & 73.5 & 77.6 & 66.4 & 56.9 \\
\midrule
\multirow[t]{2}{*}{Sonnet-3.5v1} & Code + Description & 86.6 & 86.6 & 83.4 & 86.2 & 78.5 \\
 & + Chain-of-Thought (CoT)  & 83.7 & 86.5 & 85.3 & 86.5 & 77.3 \\
\midrule
\multirow[t]{2}{*}{Sonnet-3.5v2} & Code + Description & 82.1 & 87.4 & 82.6 & 85.4 & 77.3 \\
 & + Chain-of-Thought (CoT)  & 82.2 & 88.5 & 82.3 & 85.3 & 78.9 \\
\midrule
\multirow[t]{2}{*}{LLaMA-3.1-8B} & Code + Description & 49.4 & 56.9 & 65.2 & 40.5 & 23.1 \\
 & + Chain-of-Thought (CoT)  & 55.1 & 59.3 & 63.0 & 37.6 & 25.3 \\
\midrule
\multirow[t]{2}{*}{LLaMA-3.1-70B} & Code + Description & 73.6 & 77.7 & 75.1 & 68.2 & 61.9 \\
 & + Chain-of-Thought (CoT)  & 77.7 & 76.3 & 72.4 & 65.3 & 59.8 \\
\midrule
\multirow[t]{2}{*}{LLaMA-3.1-405b} & Code + Description & 70.2 & 75.8 & 79.4 & 78.4 & 65.5 \\
 & + Chain-of-Thought (CoT)  & 72.0 & 77.0 & 80.7 & 71.5 & 66.4 \\
\midrule
\multirow[t]{2}{*}{Mistral 7B Instruct} & Code + Description & 67.1 & 46.5 & 49.6 & 48.1 & 39.3 \\
 & + Chain-of-Thought (CoT)  & 54.8 & 39.7 & 50.0 & 42.8 & 27.6 \\
\midrule
\multirow[t]{2}{*}{Mixtral 8x7B Instruct} & Code + Description & 64.9 & 51.1 & 61.1 & 47.4 & 38.4 \\
 & + Chain-of-Thought (CoT)  & 65.1 & 61.6 & 70.7 & 54.2 & 40.4 \\
\midrule
\multirow[t]{2}{*}{Mistral Small} & Code + Description & 66.0 & 54.7 & 65.8 & 50.2 & 38.3 \\
 & + Chain-of-Thought (CoT)  & 78.5 & 74.3 & 77.4 & 57.7 & 48.7 \\
\midrule
\multirow[t]{2}{*}{Mistral Large} & Code + Description & 76.9 & 70.5 & 73.5 & 78.3 & 62.0 \\
 & + Chain-of-Thought (CoT)  & 78.0 & 76.1 & 80.3 & 72.6 & 63.7 \\

\bottomrule
\end{tabular}
}
\caption{Verification accuracy (\%) across all models and two prompt configurations: using Code + Description and an additional Chain-of-Thought (CoT) step.}
\label{tab:appendix_verification_results}
\end{table*}

\subsection{Full Pipeline Results Across All Models}
Table~\ref{tab:appendix_full_pipeline_results} presents the complete pipeline results for all models using different expansion and verification methods. When analyzed alongside Table~\ref{tab:appendix_verification_results}, we can observe that the effectiveness of our verification pipeline depends on both the verification accuracy and the performance of the generation model. Models with stronger generation performance (e.g., Sonnet-3.5v1 and Haiku-3 Finetuned) require higher verification accuracy to achieve meaningful improvements, while weaker models (e.g., Haiku-3 and PLM-ICD) generally benefit from verification.

\begin{table*}[ht]
\centering
\small
\resizebox{1.8\columnwidth}{!}{%
\begin{tabular}{llrrr|r}
\toprule
 \textbf{Model} & \textbf{Expansion} & \textbf{Generation} & \textbf{+ Desc Revise} & \textbf{+ CoT Revise} & \textbf{+ Oracle} \\
\midrule
\multirow[t]{5}{*}{Haiku-3} & Siblings \(S(c)\) & 41.6 & \textbf{48.0} & 46.3 & 51.0 \\
 & Siblings \(S(c)\) + Cousins \(C(c)\) & 41.6 & 45.3 & \textbf{47.2} & 54.4 \\
 & 1-Hop Neighbors \(N_1(c)\) & 41.6 & 45.6 & \textbf{46.1} & 49.9 \\
 & 1+2-Hop Neighbors \(N_1(c) | N_2(c)\) & 41.6 & \textbf{44.5} & 43.2 & 50.1 \\
 & All Combined & 41.6 & \textbf{47.2} & \textbf{47.2} & 54.1 \\
\midrule
\multirow[t]{5}{*}{Haiku-3 (Fine-tuned)} & Siblings \(S(c)\) & 56.9 & \textbf{57.5} & 53.6 & 62.8 \\
 & Siblings \(S(c)\) + Cousins \(C(c)\) & \textbf{56.9} & 54.5 & 56.3 & 66.1 \\
 & 1-Hop Neighbors \(N_1(c)\) & \textbf{56.9} & 51.2 & 51.9 & 61.0 \\
 & 1+2-Hop Neighbors \(N_1(c) | N_2(c)\) & \textbf{56.9} & 52.4 & 51.8 & 63.6 \\
 & All Combined & 56.9 & \textbf{57.6} & 55.8 & 67.3 \\
\midrule
\multirow[t]{5}{*}{Sonnet-3.5v1} & Siblings \(S(c)\) & 55.6 & \textbf{56.5} & 54.5 & 62.3 \\
 & Siblings \(S(c)\) + Cousins \(C(c)\) & \textbf{55.6} & 51.9 & 54.9 & 65.0 \\
 & 1-Hop Neighbors \(N_1(c)\) & 55.6 & 54.6 & \textbf{56.0} & 61.6 \\
 & 1+2-Hop Neighbors \(N_1(c) | N_2(c)\) & \textbf{55.6} & 54.1 & 54.7 & 63.1 \\
 & All Combined & \textbf{55.6} & 55.5 & 55.1 & 66.4 \\
\midrule
\multirow[t]{5}{*}{PLM-ICD} & Siblings \(S(c)\) & 24.8 & \textbf{28.1} & 27.7 & 25.8 \\
 & Siblings \(S(c)\) + Cousins \(C(c)\) & 24.8 & 24.7 & \textbf{28.2} & 28.3 \\
 & 1-Hop Neighbors \(N_1(c)\) & 24.8 & 32.8 & \textbf{33.1} & 32.5 \\
 & 1+2-Hop Neighbors \(N_1(c) | N_2(c)\) & 24.8 & 28.0 & \textbf{29.8} & 30.2 \\
 & All Combined & 24.8 & 29.4 & \textbf{30.3} & 31.9 \\
\bottomrule
\end{tabular}
}
\caption{Performance of the full pipeline with different expansions and verifications (revision) across all models. The initial generations are done by the respective models, while the verification steps are done by Sonnet 3.5v2.}
\label{tab:appendix_full_pipeline_results}
\end{table*}

\subsection{Single Line Prompt}
\label{sec:single_line_prompt}

The single-line prompt from prior work~\citep{boyle2023automated}:

\begin{verbatim}
You are a professional ICD-10-CM coder.

Present your findings in a structured
JSON format as follows:
{
  "1": {
    "code": "X00.0"
  },
  "2": {
    "code": "Y00.0"
  },
  ...
}

Important: The "code" field should 
contain only the ICD-10 code itself 
(e.g., "F32.9", "I10"), without any 
descriptions or additional text.

Here is the note:
${note}
\end{verbatim}

\end{document}